\documentclass[letterpaper, 10 pt, conference]{styles/ieeeconf_new}
\IEEEoverridecommandlockouts 
\usepackage{multicol}
\usepackage[bookmarks=true]{hyperref}

\usepackage{url}
\usepackage{color}
\usepackage{wrapfig}
\usepackage{subcaption}
\newif\iffigs
\figstrue 

\usepackage[fleqn]{amsmath}
\usepackage{float}
\usepackage{setspace}
\usepackage{graphicx}
\usepackage{mathrsfs}
\usepackage{amssymb}
\usepackage{nicefrac}
\usepackage{algorithm}
\usepackage{caption}
\usepackage[noend]{algpseudocode}

\usepackage{tikz}
\usetikzlibrary{decorations.pathreplacing,calc}
\newcommand{\tikzmark}[1]{\tikz[overlay,remember picture] \node (#1) {};}
\newcommand*{\AddNote}[7]{%
    \begin{tikzpicture}[overlay, remember picture]
        \draw [decoration={brace,amplitude=0.5em},decorate,ultra thick,black]
            ($(#3)!(#1.north)!($(#3)-(0,1)$)$) --  
            ($(#3)!(#2.south)!($(#3)-(0,1)$)$)
                node [align=center, text width=#7cm, pos=#6, anchor=#5, font=\bfseries] {#4};
    \end{tikzpicture}
}

\usepackage{flushend}

\makeatletter
\newcommand\fs@spaceruled{\def\@fs@cfont{\bfseries}\let\@fs@capt\floatc@ruled
  \def\@fs@pre{\vspace{0.4\baselineskip}\hrule height.8pt depth0pt \kern2pt}%
  \def\@fs@post{\vspace{-0.4\baselineskip}\kern2pt\hrule\relax\vspace{-12pt}}%
  \def\@fs@mid{\kern2pt\hrule\kern2pt}%
  \let\@fs@iftopcapt\iftrue}
\makeatother

\usepackage{lipsum}
\newcommand\myeq{\mkern1.5mu{=}\mkern1.5mu}
\newcommand{\customSpacing}[1]{\setlength{\belowdisplayskip}{#1} \setlength{\belowdisplayshortskip}{#1}\setlength{\abovedisplayskip}{#1} \setlength{\abovedisplayshortskip}{#1}}
\usepackage{xfrac}

\title{\bfseries MPCGPU: Real-Time Nonlinear Model Predictive Control\\through Preconditioned Conjugate Gradient on the GPU}

\author{Emre Adabag$^{1}$, Miloni Atal$^{1*}$, William Gerard$^{1*}$, Brian Plancher$^{2}$
\thanks{This material is based upon work supported by the National Science Foundation (under Award 2246022). Any opinions, findings, conclusions, or recommendations expressed in this material are those of the authors and do not necessarily reflect those of the funding organizations.}
\thanks{$^{1}$Emre Adabag, Miloni Atal, and William Gerard are with the School of Engineering and Applied Science, Columbia University, New York, NY. {\tt\footnotesize \{ea2944, ma4338, wg2404\}@columbia.edu}}%
\thanks{$^{2}$Brian Plancher is with Barnard College, Columbia University, New York, NY. {\tt\footnotesize bplancher@barnard.edu}}%
\thanks{$^{*}$These authors contributed equally to this work.}%
}

\begin{document}

\maketitle
\thispagestyle{empty}
\pagestyle{empty}


\begin{abstract}
    Nonlinear Model Predictive Control (NMPC) is a state-of-the-art approach for locomotion and manipulation which leverages trajectory optimization at each control step. While the performance of this approach is computationally bounded, implementations of direct trajectory optimization that use iterative methods to solve the underlying moderately-large and sparse linear systems, are a natural fit for parallel hardware acceleration.
In this work, we introduce MPCGPU, a GPU-accelerated, real-time NMPC solver that leverages an accelerated preconditioned conjugate gradient (PCG) linear system solver at its core.
We show that MPCGPU increases the scalability and real-time performance of NMPC, solving larger problems, at faster rates. In particular, for tracking tasks using the Kuka IIWA manipulator, MPCGPU is able to scale to kilohertz control rates with trajectories as long as 512 knot points.
This is driven by a custom PCG solver which outperforms state-of-the-art, CPU-based, linear system solvers by at least 10x for a majority of solves and 3.6x on average.
\end{abstract}

\section{Introduction} \label{sec:intro}
Nonlinear Model Predictive Control (NMPC) is a feedback control strategy which repeatedly solves finite horizon optimal control problems (OCP) in real time, enabling robots to adapt to changes in their environment. This approach has seen great recent success in applications to both locomotion and manipulation~\cite{hogan2018reactive,sleiman2021unified,tranzatto2022cerberus,wensing2022optimization,Kuindersma23Talk}. 

Most implementations of NMPC leverage trajectory optimization~\cite{Betts01} to solve the underlying optimal control problems. Two popular classes of these algorithms are shooting methods and direct methods. Shooting methods parameterize only the input trajectory and use Bellman's optimality principle \cite{Bellman57} to iteratively solve a sequence of smaller optimization problems~\cite{Mayne66,Jacobson70}. Direct methods explicitly represent the states, controls, dynamics, and any additional constraints, leading to moderately-large nonlinear programs with structured sparsity patterns~\cite{Nocedal06}.

There has been historical interest in parallel strategies~\cite{Betts91} for solving trajectory optimization problems. This is growing increasingly important with the impending end of Moore's Law and the end of Dennard Scaling, which have led to a utilization wall that limits the performance a single CPU chip can deliver~\cite{Esmaeilzadeh11,Venkatesh10}. Several more recent efforts have shown that significant computational benefits are possible by exploiting the natural parallelism in the computation of the (gradients of the) dynamics and cost functions on GPUs and FPGAs~\cite{Antony17,Pan19,Plancher21,Neuman21,Plancher22GRiDGPUAcceleratedRigidBodyDynamicsAnalyticalGradientsb,Lee22ParallelILQR,Neuman23RoboShape,yang2023rbdcore}. However, multiple-shooting and consensus approaches to computing trajectory updates at each algorithmic iteration~\cite{Giftthaler17,Farshidian17,Kouzoupis16,jiang2020parallel} have only seen modest gains when implemented on alternative hardware platforms~\cite{Plancher18,Plancher19a}.

On the other hand, direct methods naturally expose more parallelism that can be exploited through hardware acceleration. Importantly, these approaches are computationally dominated by the solutions of a moderately-large and sparse linear systems. Iterative methods, like the Preconditioned Conjugate Gradient (PCG) algorithm~\cite{eisenstat1981efficient}, are particularly well-suited for parallel solutions of linear systems, as they are computationally dominated by matrix-vector products and vector reductions~\cite{Saad03,Plancher22Dissertation}, and have shown past success in outperforming state-of-the-art CPU implementations for solving very-large linear systems on GPUs~\cite{Helfenstein12ParallelpreconditionedconjugategradientalgorithmGPU,Schubiger20}. 

\iffigs
\begin{figure}[!t]
    \centering
    \includegraphics[width=0.71\columnwidth]{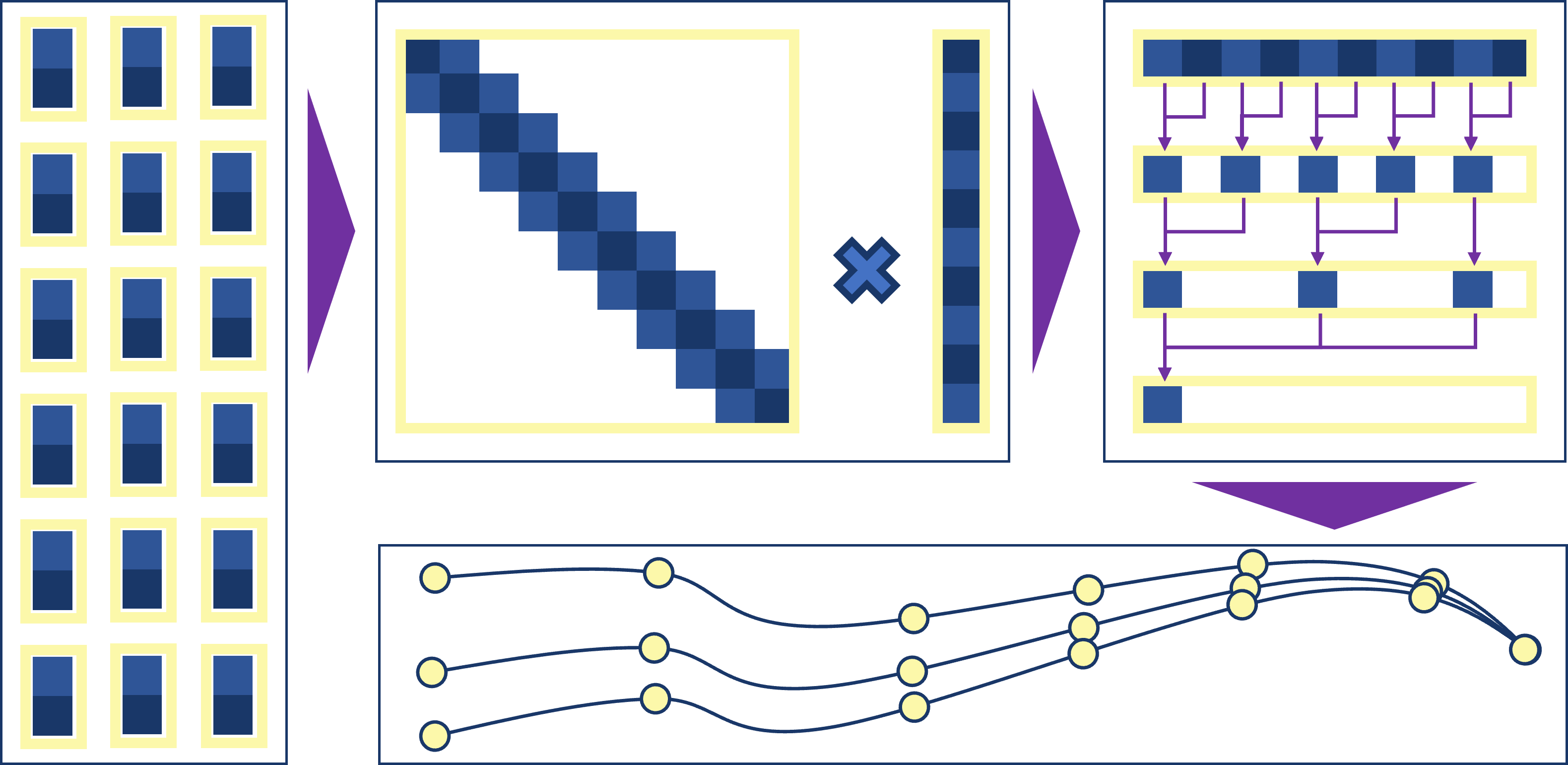}
    \vspace{2pt}
    \caption{At each control step, MPCGPU uses a parallel construction of the Schur complement of the Karush-Kuhn-Tucker (KKT) system, a parallel-friendly PCG solver, and a parallel line search to provide real-time performance through GPU acceleration.}
    \label{fig:mpcgpu_overview_small}
    \vspace{-18pt}
\end{figure}
\fi

In this work, we introduce MPCGPU, a GPU-accelerated, real-time NMPC solver that exploits the structured sparsity and the natural parallelism in direct trajectory optimization (see Figure~\ref{fig:mpcgpu_overview_small}). At our solver's core is a custom, accelerated implementation of PCG tuned for the Schur complement of the KKT systems of trajectory optimization problems.

We show that MPCGPU increases the scalability and real-time performance of NMPC, solving larger problems, at faster rates. In particular, for tracking tasks using the Kuka IIWA manipulator, MPCGPU is able to scale to kilohertz rates with trajectories as long as 512 knot points. This is driven by a custom, GPU-accelerated, PCG solver which outperforms state-of-the-art, CPU-based, linear system solvers by at least 10x for a majority of solves and 3.6x on average.
We release our software and experiments open source 
at: {\small \href{https://github.com/a2r-lab/MPCGPU}{\color{blue} \texttt{https://github.com/a2r-lab/MPCGPU}}}.

\section{Related Work} \label{sec:related}
There has been a significant amount of prior work developing general purpose sparse linear system solvers on the GPU 
both using factorization-based approaches~\cite{Jung06,Yang12,Venetis15,Hogg16,Hu17,Yeralan17,swirydowicz2022linear,cole2023exploiting,shin2023accelerating,pacaud2023accelerating}, as well as iterative methods~\cite{Saad03,Schubiger20,Bolz03,Liu13,Anzt17,Anzt18,Flegar19,tiwari2022strategies}. There has also been work developing and implementing Block-Cyclic-Reduction and other tree-structured methods that are optimized for block-tridiagonal systems~\cite{Chang12,Chang14,Dieguez15,Lamas18}. These general purpose approaches have found speedups through GPU usage, but only once the problem size grows to more than tens if not hundreds of thousands of variables or for instances of (linear) (power-flow) problems~\cite{Schubiger20,swirydowicz2022linear,cole2023exploiting,shin2023accelerating,pacaud2023accelerating,naumov2011incomplete}.
As such, these general purpose solvers are not performant for most trajectory optimization problems (e.g., our examples in Section~\ref{sec:results} have 448 to 7,168 variables). 

For the nonlinear trajectory optimization problem, evolutionary, particle-swarm, Monte-Carlo, and other sampling based approaches have been implemented on GPUs~\cite{Heinrich15,Wu16,Williams17,Phung17,Hyatt17,Ohyama17,Rathai19,Wang19a,Hyatt20}. Most prior work on gradient-based parallel nonlinear trajectory optimization has been fully confined to the CPU~\cite{Frasch13,Kouzoupis16,Giftthaler17,Farshidian17,jiang2020parallel,Astudillo22}, relied on the CPU for many of the computations~\cite{Gang12,Gade12}, focused only on the problem of optimizing BLAS functions on the GPU~\cite{Huang11,Yu17}, or was limited to GPU acceleration of the naturally parallel (gradients of the) dynamics and cost functions~\cite{Antony17,Plancher21,Plancher22GRiDGPUAcceleratedRigidBodyDynamicsAnalyticalGradientsb,Lee22ParallelILQR}. 

There are two existing lines of work which fully implemented gradient-based nonlinear trajectory optimization on the GPU. The first leveraged shooting based methods and found them to not expose much natural parallelism, limiting their performance~\cite{Plancher18,Plancher19a}. The second used a Block-Cyclic-Reduction-based direct method to exploit the particular structure exposed by position-based dynamics~\cite{Pan19}. 

This work adds to the literature by designing a GPU-accelerated, gradient-based, direct trajectory optimization solver for standard reduced-coordinate dynamics~\cite{Featherstone08} leveraging a custom parallel PCG solver at its core. 

\section{Background} \label{sec:background}
\subsection{Direct Trajectory Optimization}
\label{sec:background_trajopt}
Trajectory optimization~\cite{Betts01}, also known as numerical optimal control, solves an (often) nonlinear optimization problem to compute a robot's path through an environment as a series of states $X\myeq \{x_0,\dotsi,x_N\}$ and controls $U\myeq  \{u_0,\dotsi,u_{N-1}\}$ for $x$ $\in \mathbb{R}^n$ and $u$ $\in \mathbb{R}^m$. These problems model the robot as a discrete-time dynamical system,
{\customSpacing{5pt}
\begin{equation}
x_{k+1} = f(x_k,u_k,h), \quad x_0 = x_s, \label{eq:dynamics}
\end{equation}
with a timestep $h$, and minimize an additive cost function,
\begin{equation}
    J(X,U) = \ell_f(x_N) + \sum^{N-1}_{k = 0} \ell(x_k,u_k). \label{eq:cost}
\end{equation}
}
Direct methods for trajectory optimization form a moderately-large and sparse nonlinear program. While there are a variety of algorithmic approaches used to solve these problems, most methods can be reduced to a three step process which is repeated until convergence~\cite{Nocedal06, Wachter06, Gill05}.

\emph{Step 1:}
Compute a second-order Taylor expansion of our problem along a nominal trajectory, resulting in the following quadratic program (QP) for the deviation ($\delta X$, $\delta U$):
\begin{equation}
\begin{split}
    \min_{\substack{\delta X, \delta U}} 
        &\;\; \tfrac{1}{2}\delta x_N^T Q_N \delta x_N + q_N^T \delta x_N + \\ 
        \sum_{k=0}^{N-1} &\tfrac{1}{2}\delta x_k^T Q \delta x_k + q^T \delta x_k + \tfrac{1}{2}\delta u_k^T R \delta u_k + r^T \delta u_k\\
    \text{s.t.\quad} &\delta x_0 = x_s - x_0, \\
    \delta x&_{k+1} - A_k \delta x_k - B_k \delta u_k = f(x_k,u_k) - x_{k+1}\\
        & \quad \quad \quad \quad \quad \quad \quad \quad \quad \quad \quad \quad \forall k \in \mathbb{Z} \cap [0,N)
\end{split}
\end{equation}

\emph{Step 2:} Compute $\delta X^*, \delta U^*$ by solving the KKT system:
\begin{equation} \label{eq:KKT}
\begin{split}
    &\begin{bmatrix} G & C^T \\ C & 0 \end{bmatrix} \begin{bmatrix} -\delta Z \\ \lambda \end{bmatrix} = \begin{bmatrix} g \\ c \end{bmatrix}\\
\end{split}
\end{equation}

where:
\begin{equation*} \label{eq:KKTSimplifications}
\scalebox{0.99}{$
\begin{split}
  \delta z_k &= \begin{bmatrix} \delta x_k & \delta u_k\end{bmatrix}^T \quad \quad \quad \delta z_N = \delta x_N\\
    G &= \begin{bmatrix}
        Q_0 & & & \\
        & R_0 & & \\
        & & \ddots & \\
        & & & Q_N \\
        \end{bmatrix} \\
    g &= \begin{bmatrix} q_0 & r_0 & q_1 & r_1 & \dots & q_N \end{bmatrix}^T \\
    C &= \begin{bmatrix}
           I &      &        &          &          & \\
        -A_0 & -B_0 & I      &          &          & \\
             &      & \ddots & -A_{N-1} & -B_{N-1} & I \\
        \end{bmatrix} \\
    e_k &= x_{k+1} - f(x_k,u_k)\\
    c &= \begin{bmatrix} x_0 - x_s & e_0 & e_1 & \dots & e_{N-1} \end{bmatrix}^T
\end{split}
$}
\end{equation*}
    
\emph{Step 3:} Apply the update step, $\delta X^*, \delta U^*$, while ensuring descent on the original nonlinear problem through the use of a merit-function and a trust-region or line-search~\cite{Nocedal06}.


\subsection{The Schur Complement Method}
\label{sec:background_schur}
One approach to solving Equation~\ref{eq:KKT} is through a two step process which forms the symmetric positive definite \emph{Schur Complement}, $S$, and first solves for $\lambda^*$ and then $\delta z^*$: 
\begin{equation} \label{eq:schur}
\begin{split}
    S &= -CG^{-1}C^T        \hspace{20pt}      
    \gamma = c - CG^{-1}g \\
    S\lambda^* &= \gamma  \hspace{54pt}     
    \delta z^* = -G^{-1}(g - C^T\lambda^*). \\
\end{split}
\end{equation}
By defining the variables $\theta$, $\phi$, and $\zeta$,
\begin{equation} \label{eq:trajoptSchur_1}
\begin{split}
    \theta_k &= A_k Q_k^{-1} A_k^T + B_k R_k^{-1} B_k^T + Q_{k+1}^{-1} \\
    \phi_k &= -A_k Q_k^{-1} \\
    \zeta_k &= -A_k Q_k^{-1} q_k - B_k R_k^{-1} r_k + Q_{k+1}^{-1} q_{k+1},
\end{split}
\end{equation}
$S$ and $\gamma$ take the form:
\begin{equation} \label{eq:trajoptSchur_2}
\begin{split}
    S &= -\begin{bmatrix}
        Q_0^{-1} & \phi_0^T & & & & \\
        \phi_0 & \theta_0 & \phi_1^T & & & \\
        & & \ddots & \phi_{N-2} & \theta_{N-2} & \phi_{N-1}^T\\
        & & & & \phi_{N-1} & \theta_{N-1}\\
    \end{bmatrix} \\
    \gamma &= c - \begin{bmatrix} Q_0^{-1}q_0 & \zeta_0 & \zeta_1 & \dots & \zeta_{N-1} \end{bmatrix}^T
\end{split}
\end{equation}

\iffigs
\begin{figure*}[!h]
    \centering
    \vspace{5pt}
    \includegraphics[width=1.9\columnwidth]{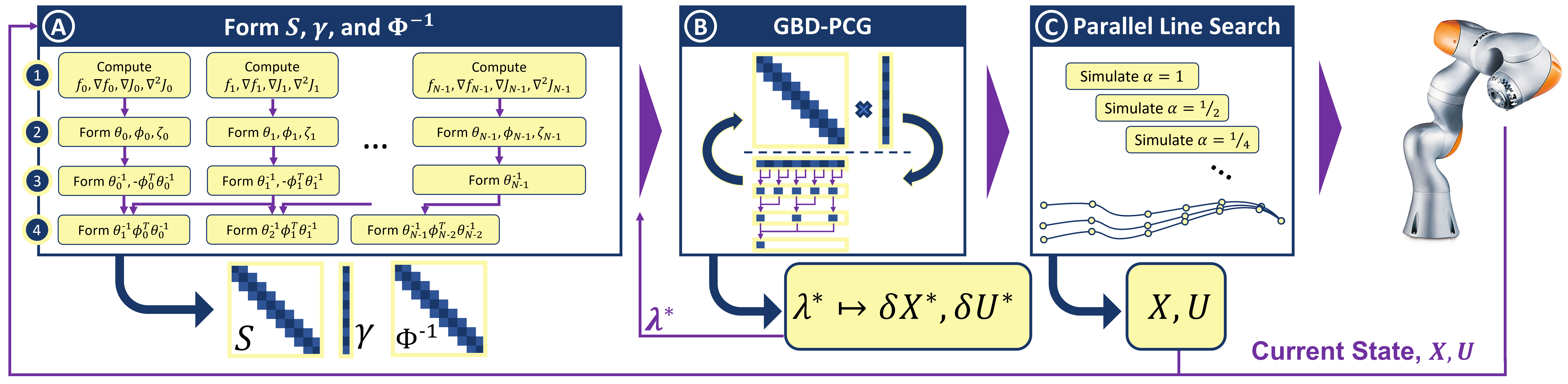}
    \vspace{-2pt}
    \caption{A high level overview of MPCGPU which: 1) in parallel on the GPU computes $S$, $\gamma$, and $\Phi^{-1}$ and stores those values in an optimized dense format, 2) uses our GBD-PCG solver to compute $\lambda^*$ and reconstructs $\delta X^*, \delta U^*$ through GPU-friendly matrix-vector multiplications and vector reductions, and 3) leverages a parallel line search to compute the final trajectory, $X, U$. This trajectory is then passed to the (simulated) robot and the current state of the (simulated) robot is measured and fed back into our solver which is run again, warm-started with our last solution.}
    \label{fig:mpcgpu_overview}
    \vspace{-14pt}
\end{figure*}
\fi

\subsection{Iterative Methods}
\label{sec:background_krylov}
Iterative methods solve the problem $S\lambda^* = \gamma$ for a given $S$ and $\gamma$ by iteratively refining an estimate for $\lambda$ up to some tolerance $\epsilon$. The most popular of these methods is the conjugate gradient (CG) algorithm which has been used for state-of-the-art results on large-scale optimziation problems on the GPU~\cite{naumov2011incomplete,Schubiger20GPUAccelerationADMMLargeScaleQuadraticProgramming}.
The convergence rate of CG is directly related to the spread of the eigenvalues of $S$.
Thus, a preconditoning matrix $\Phi \approx S$ is often applied to instead solve the equivalent problem with better numerical properties: $\Phi^{-1} S\lambda^* = \Phi^{-1} \gamma$. 
To do so, the preconditioned conjugate gradient (PCG) algorithm leverages matrix-vector products with $S$ and $\Phi^{-1}$, as well as vector reductions, both parallel friendly operations (see Algorithm~\ref{alg:PCG}).

\subsection{Graphics Processing Units (GPUs)}
\label{sec:background_gpu}
Compared to a multi-core CPU, a GPU is a larger set of simpler processors, optimized for parallel execution of identical instructions. 
GPUs are best at computing regular and separable computations, over large data sets, with limited synchronization (e.g., large matrix multiplication)~\cite{Nickolls08ScalableParallelProgrammingCUDA}.
Our work uses NVIDIA’s CUDA extensions to C++~\cite{NVIDIA18}.

\section{The MPCGPU Solver} \label{sec:features}
In this section we describe the design of the MPCGPU solver which exploits the sparsity and natural parallelism found in direct trajectory optimization algorithms and iterative linear system solvers.
To further promote efficient GPU acceleration, 
unlike generic approaches, which require a kernel launch and CPU-GPU synchronization for each matrix operation, MPCGPU uses only three kernels that are asynchronously queued, resulting in only a single CPU-GPU synchronization. We also only transfer the initial and final values between the CPU and GPU to reduce I/O overheads.

As shown in Figure~\ref{fig:mpcgpu_overview}, our approach can be broken down into a three step process. At each control step we first compute each block row of the Schur complement system, $S$ and $\gamma$, as well as our preconditioner, $\Phi^{-1}$, in parallel by taking advantage of the structured sparsity of those matrices. Next, we use our custom GPU-optimized, warm-started, PCG solver, GBD-PCG (Algorithm~\ref{alg:PPCG}), to compute the optimal Lagrange multipliers, $\lambda^*$, and reconstruct the optimal trajectory update, $\delta X^*, \delta U^*$. Finally we leverage a parallel line search to compute the final trajectory $X, U$ which we send to the (simulated) robot for execution and simultaneously measure the current state of the (simulated) robot to begin our next control step. In the remainder of this section we provide further details on our approach. Our open-source implementation can be found at: 
{\small \href{https://github.com/a2r-lab/MPCGPU}{\color{blue} \texttt{https://github.com/a2r-lab/MPCGPU}}}.

\floatstyle{spaceruled}
\restylefloat{algorithm}
\begin{algorithm}[!t]
\begin{spacing}{1.2}
\begin{algorithmic}[1]
\caption{Preconditioned Conjugate Gradient (PCG) \newline ($S,\Phi^{-1},\gamma,\lambda,\epsilon$) $\rightarrow$ $\lambda^*$} \label{alg:PCG}
    \State $r = \gamma - S\lambda$ \tikzmark{topI}
    \State $\tilde{r}, p = \Phi^{-1} r$ \quad \quad \quad \quad \quad \quad \quad \quad \; \tikzmark{sideI}
    \State $\eta = r^T \tilde{r}$ \tikzmark{bottomI}
    \For{iter $i = 1:\text{max\_iter}$} \tikzmark{topM}
        \State $\alpha = \eta / (p^TSp)$
        \State $r = r - \alpha Sp$
        \State $\lambda = \lambda + \alpha p$
        \State $\tilde{r} = \Phi^{-1} r$
        \State $\eta^\prime = r^{T} \tilde{r}$
        \If{$\eta^\prime < \epsilon$} return $\lambda$ \; \quad \quad \tikzmark{sideM}
        \EndIf
        \State $\beta = \eta^\prime / \eta$
        \State $p = \tilde{r} + \beta p$
        \State $\eta = \eta^\prime$ \tikzmark{bottomM}
    \EndFor
    \State return $\lambda$
\end{algorithmic}
\end{spacing}
\AddNote{topI}{bottomI}{sideI}{Initialization}{west}{0.5}{2.5}
\AddNote{topM}{bottomM}{sideM}{Main Loop}{west}{0.5}{2.25}
\vspace{-10pt}
\end{algorithm}

\subsection{Parallel Computation of $S, \gamma$, and $\Phi^{-1}$}

To efficiently compute $S, \gamma$, and $\Phi^{-1}$ on the GPU, we need to find a naturally parallel approach to form the values as well as an efficient data storage format that minimizes overheads. We also need to find an effective preconditioner that is parallel-friendly in its computation. 

We first leverage the block-tridiagonal structure of the Schur complement, $S$, as shown in Equation~\ref{eq:trajoptSchur_2}, which is for the most part independent across timesteps, $k$, for each block-row. This pattern also extends to each block-row of the $\gamma$ vector. To further remove the need for synchronizations, for each $k$, we also compute the only cross-timestep quantities, $Q_{k+1}$ and $q_{k+1}$. While this results in those terms being computed twice, it still proves to be more efficient than forcing a synchronization point between all block-rows. To ensure efficient computation of the underlying dynamics and kinematic quantities, we leverage the GRiD library, which was shown to outperform state-of-the-art CPU libraries even when taking into account I/O overheads~\cite{Plancher22GRiDGPUAcceleratedRigidBodyDynamicsAnalyticalGradientsb}.

We further parallelize across and within the many small matrix inversions and matrix multiplications within each parallel block-row computation. Leveraging best practices~\cite{Plancher21,Plancher22GRiDGPUAcceleratedRigidBodyDynamicsAnalyticalGradientsb}, we also group together the various types of mathematical operations, storing intermediate values in shared memory, and re-ordering computations where needed.

We leverage the Symmetric Stair Preconditioner~\cite{Plancher22Dissertation,bu2024symmetric} which is a parallel-friendly preconditioner optimized for block-tridiagonal systems which has an analytical inverse and results in the block-tridiagonal matrix:
\begin{equation}
    \Phi^{-1} = \begin{bmatrix}
        Q_0 & -Q_0 \phi_0^T \theta_0^{-1} &  \\
        -\theta_0^{-1} \phi_0 Q_0 & \;\; \theta_0^{-1} \;\; & -\theta_0^{-1} \phi_1^T \theta_1^{-1} \\
         & -\theta_1^{-1} \phi_1 \theta_0^{-1} & \theta_1^{-1} \\
        & & \ddots \\
    \end{bmatrix}\\
\end{equation}
The structure of $\Phi^{-1}$ also permits mostly parallel computation as only the values of each $\theta_k^{-1}$ need to be shared across timesteps.
Our approach thus requires only a single global synchronization across blocks, and allows us to store the block-tridiagonal $S$ and $\Phi^{-1}$ matrices in a custom, compressed, dense format for increased IO bandwidth and memory efficiency.

\subsection{GPU Parallel PCG for Block-Tridiagonal Systems} \label{sec:GPUPCG}

The core of our solver is a custom GPU Parallel PCG implementation specifically optimized for block-tridiagonal systems, GBD-PCG (Algorithm~\ref{alg:PPCG}). That is, we leverage the sparsity structure of $S$ and $\Phi^{-1}$ to maximize cache usage and natural parallelism resulting in a refactored, low-latency implementation with minimal synchronizations. These optimizations can be leveraged in the most computationally expensive part of the algorithm, the large matrix-vector products in lines 5, 6, and 8 of Algorithm~\ref{alg:PCG}, as each element of the product depends on at most $3n_b$ elements from the matrix and $3n_b$ elements from the vector, where $n_b$ is the block dimension. We exploit this by grouping threads that access similar elements into thread blocks and storing $S$, $\Phi^{-1}$, and all PCG iterates concurrently in shared (cache) memory on the GPU. We also operate as many steps of the algorithm fully in parallel as possible between the thread synchronizations needed for the parallel reductions of scalar values on lines 6, 12, and 21 of Algorithm~\ref{alg:PPCG}. Similarly, we only use device memory (RAM) for those scalar reductions and for the values of $p$ and $r$ that need to be shared between blocks on lines 9 and 18 of Algorithm~\ref{alg:PPCG}. This means that the choice of a sparse preconditioner not only enables its efficient computation and memory storage, but also reduces the number of synchronizations and amount of memory that needs to be shared through RAM during each PCG iterate. This holistic co-design across algorithm stages is part of the reason why MPCGPU is so performant. Finally, we warm-start the values for $\lambda$ based on the previous solve which we found greatly increased overall performance by reducing the number of PCG iterations needed for convergence.

\floatstyle{spaceruled}
\restylefloat{algorithm}
\begin{algorithm}[!t]
\begin{spacing}{1.2}
\begin{algorithmic}[1]
\caption{GPU Parallel PCG for Block-Tridiagonal Systems (GBD-PCG) ($S,\Phi^{-1},\gamma,\lambda,\epsilon$) $\rightarrow$ $\lambda^*$} \label{alg:PPCG}
    \For{block $b = 0:N$ \textbf{in parallel}} \tikzmark{topI}
        \State $r_b = \gamma_b - S_{b} \lambda_{b-1 : b+1}$
        \State Load $r_{b-1}$, $r_{b+1}$
        \State $\tilde{r}_b,p_b = \Phi^{-1}_{b} r_{b-1 : b+1}$ \quad \quad \quad \quad \quad \quad \; \tikzmark{sideI}
        \State $\eta_b = r_b^T\tilde{r_b}$
    \EndFor
    \State $\eta = \textbf{\text{ParallelReduce}}(\eta_b)$ \tikzmark{bottomI}
    \For{iter $i = 1:\text{max\_iter}$} \tikzmark{topM}
        \For{block $b = 0:N$ \textbf{in parallel}}
            \State Load $p_{b-1}$, $p_{b+1}$ 
            \State $\Upsilon_b = S_{b} p_{b-1 : b+1}$
            \State $\upsilon_b = p_b \Upsilon_b$
        \EndFor
        \State $\upsilon = \textbf{\text{ParallelReduce}}(\upsilon_b)$
        \For{block $b = 0:N$ \textbf{in parallel}}
            \State $\alpha = \eta / \upsilon$
            \State $\lambda_b = \lambda_b + \alpha p_b$
            \State $r_b = r_b - \alpha \Upsilon_b$
        \EndFor
        \For{block $b = 0:N$ \textbf{in parallel}}
            \State Load $r_{b-1}$, $r_{b+1}$ \quad \quad \quad \quad \quad \quad \;\;\; \tikzmark{sideM}
            \State $\tilde{r}_b = \Phi^{-1}_b r_{b-1:b+1}$
            \State $\eta^\prime_b = r^{T}_b \tilde{r}_b$
        \EndFor
        \State $\eta^\prime = \textbf{\text{ParallelReduce}}(\eta^\prime_b)$
        \If{$\eta^\prime < \epsilon$} return $\lambda$
        \EndIf
        \For{block $b = 0:N$ \textbf{in parallel}}
            \State $\beta = \eta^\prime / \eta$
            \State $p_b = \tilde{r}^\prime_b + \beta p_b$
            \State $\eta = \eta^\prime$ \tikzmark{bottomM}
        \EndFor
    \EndFor
    \State return $\lambda$
\end{algorithmic}
\end{spacing}
\AddNote{topI}{bottomI}{sideI}{Initiali-\\zation}{west}{0.5}{1.5}
\AddNote{topM}{bottomM}{sideM}{Main\\Loop}{west}{0.5}{1.25}
\vspace{-10pt}
\end{algorithm}

\subsection{Parallel Line Search}
We leverage a parallel line search, computing all possible iterates for $\alpha \in \mathbb{A}$ in parallel\footnote{In this work we use $\mathbb{A} = \left\{1,\frac{1}{2},\dotsi,\frac{1}{256}\right\}$, but any decaying set of fractional values can be used in practice.} and selecting the iterate with the best value according to its L1 merit function~\cite{Nocedal06}.
This allows MPCGPU to evaluate all possible line search iterates in the same amount of time as it would take to compute a single iterate under a standard backtracking approach. 
Importantly, this not only reduces latency of this step, 
but has also been shown to improve the convergence of NMPC on similar whole-body trajectory tracking problems~\cite{Plancher18}.

\section{Results} \label{sec:results}
In this section we present a two-part evaluation of MPCGPU through a case study of online, dynamic, multi-goal, end-effector position tracking using whole-body NMPC for a simulated Kuka IIWA manipulator. First, we compare the performance of our underlying GBD-PCG iterative linear system solver with the state-of-the-art, CPU-based, QDLDL solver~\cite{stellato20osqp}. Second, we show how the end-to-end performance enabled by MPCGPU allows us to scale to long time horizons and fast control rates. Source code accompanying this evaluation can be found at {\small \href{https://github.com/a2r-lab/MPCGPU}{\color{blue} \texttt{https://github.com/a2r-lab/MPCGPU}}}.

\subsection{Methodology}
Results were collected on high-performance workstation with a $3.2$GHz 16-core Intel i9-12900K and a $2.2$GHz NVIDIA GeForce RTX 4090 GPU running Ubuntu 22.04 and CUDA 12.1. Code was compiled with \texttt{g++11.4}, and time was measured with the Linux system call \texttt{clock\_gettime()}, using \texttt{CLOCK\_MONOTONIC} as the source. 
Our performance analysis is drawn from 100 NMPC trials of a 10 second, 5 goal, pick-and-place circuit for a simulated Kuka IIWA-14 (see Figure~\ref{fig:task}).
Each NMPC trial consists of thousands of linear system solves which are needed for the many iterations of the underlying trajectory optimization problem for end-effector position tracking solved at each control step.
All hyperparameter values can be found in our open-source source code and resulted in an average tracking error of $\sim$10cm, providing similar performance as previous experiments with GPU accelerated NMPC~\cite{Plancher19a}. In particular, we note that all solvers used the same quadratic cost functions for each amount of knot points, and solver-specific hyperparameter values were independently tuned for maximal performance.

\subsection{Linear System Solver Performance} \label{sec:results:linear}
We evaluate the performance of our underlying GBD-PCG linear system solver over its thousands of solves during each of our 100 NMPC trials running at a 500hz control rate and compare its performance to the state-of-the-art CPU-based QDLDL solver~\cite{stellato20osqp} operating in the same context.

\textbf{Average Solve Time:} Our results show that our GPU based solver outperforms QDLDL across most problem sizes and is only marginally slower at the smallest problem size, obtaining as much as a 3.6x average speedup (see Figure~\ref{fig:solve-times}). This is driven by the GPU's ability to leverage large scale parallelism to gracefully scale to larger problem sizes. We note that the speedup plateaus at 256 knot points as we begin to run out of hardware resources on our specific GPU. These results show that unlike generic approaches that become performant at tens to hundreds of thousands of variables~\cite{Schubiger20,naumov2011incomplete}, our domain-specific co-design approach enables the GPU to outperform the CPU even on moderately-sized linear systems (our experiments range from 448 to 7,168 variables). 

\iffigs
\begin{figure}[!t]
    \centering
    \vspace{10pt}
    \includegraphics[width=0.99\columnwidth]{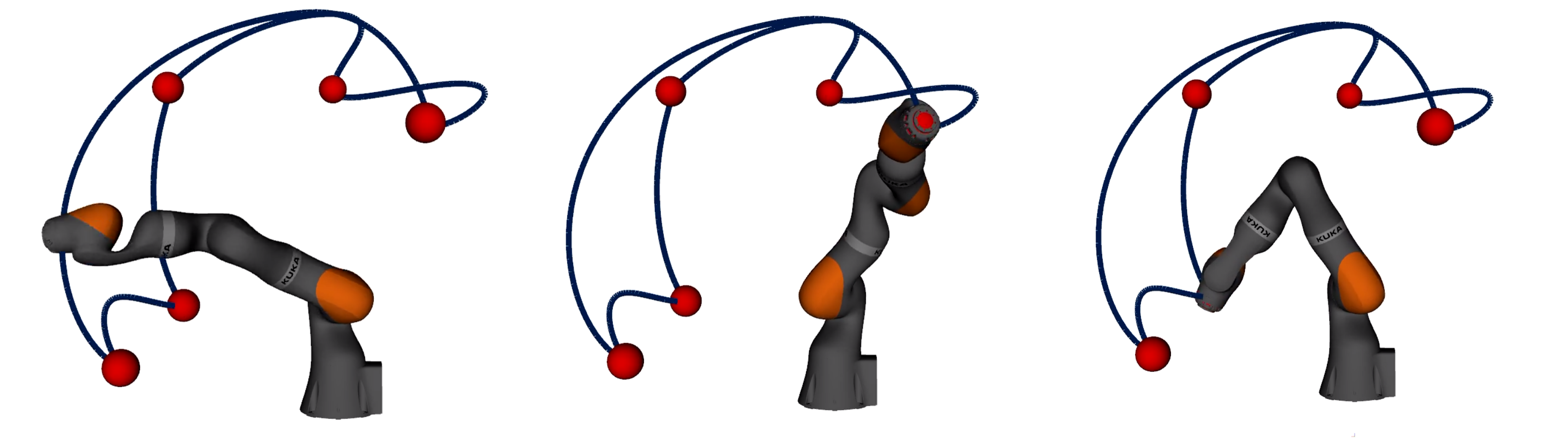}
    \vspace{-12pt}
    \caption{Screenshots from the 5 goal, pick-and-place circuit used in our experiments.}
    \label{fig:task}
\end{figure}
\fi

\iffigs
\begin{figure}[!t]
    \centering
    \includegraphics[width=0.99\columnwidth]{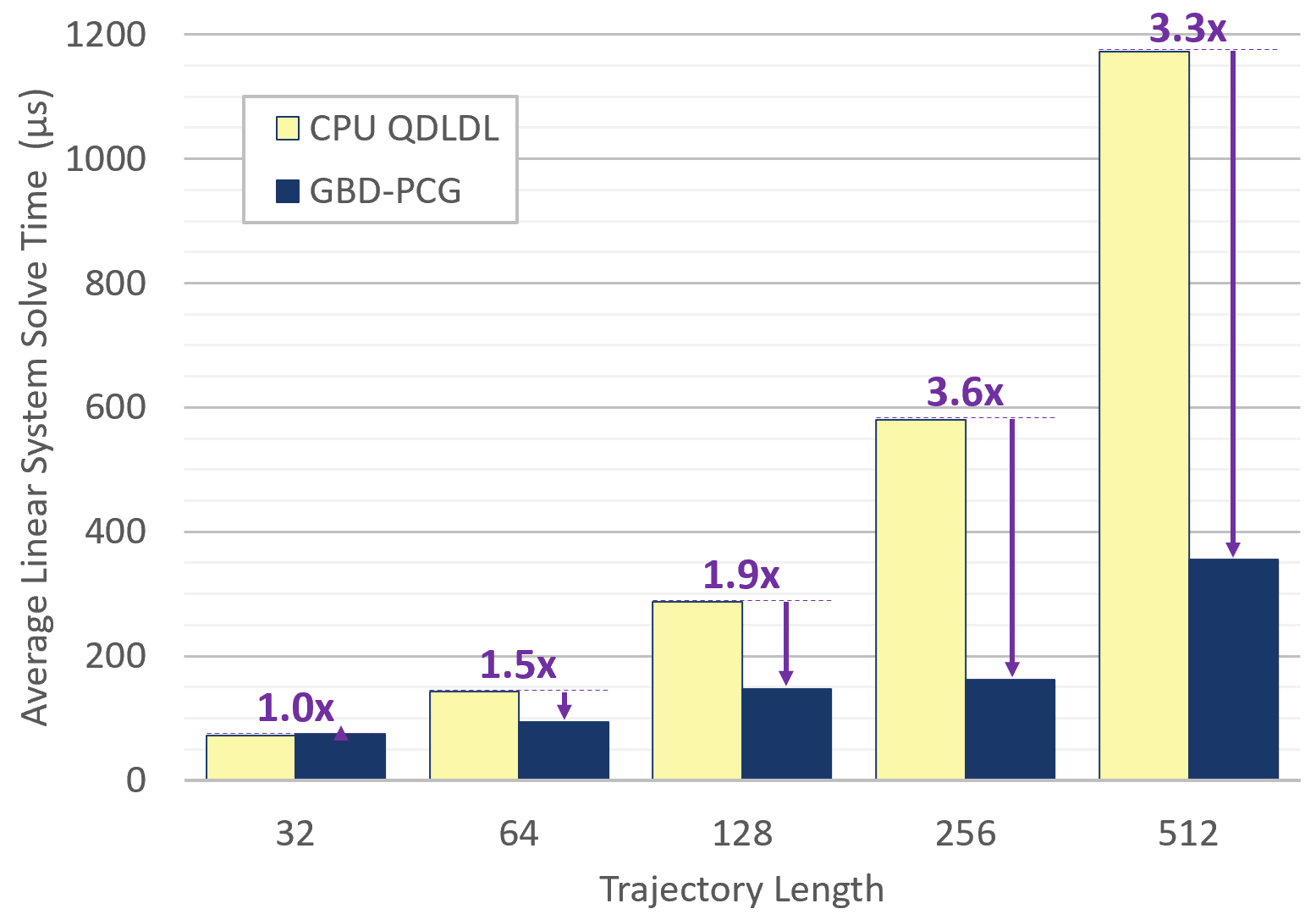}
    \vspace{-15pt}
    \caption{Average linear system solve time for the thousands of solves within each of the 100 iterations of NMPC. As the problem scales, so does the advantage of GBD-PCG over QDLDL achieving up to a 3.6x average speedup.}
    \label{fig:solve-times}
\end{figure}
\fi

\iffigs
\begin{figure}[!t]
    \centering
    \includegraphics[width=0.99\columnwidth]{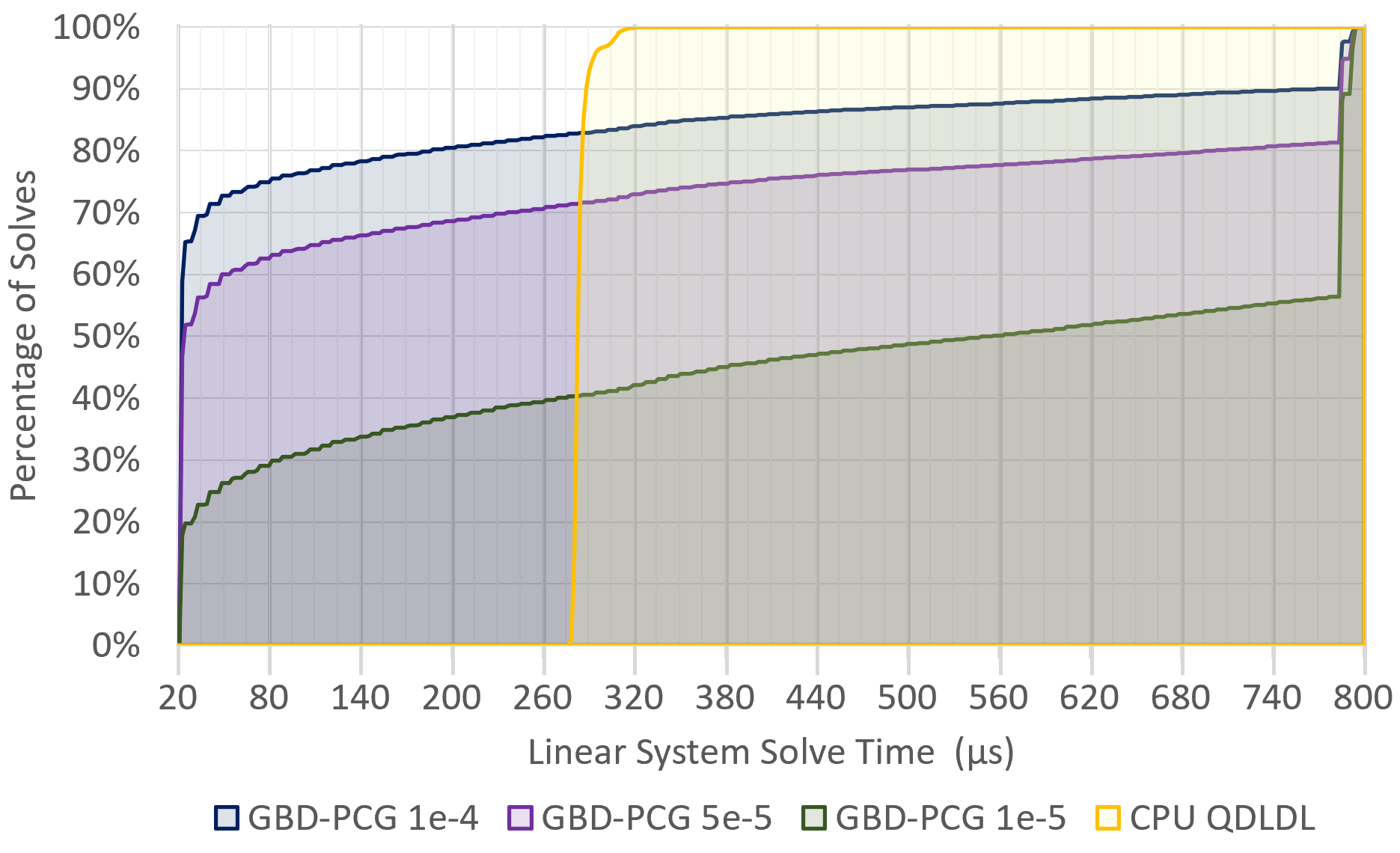}
    \caption{Cumulative density function of the linear system solve times for a trajectory length of 128 for QDLDL and GBD-PCG under multiple different exit tolerances, $\epsilon$. We note that the bi-modal distribution for GBD-PCG is often much faster than the uni-modal distribution for QDLDL. For example, for $\epsilon = 1e^{-4}$, 65\% of GBD-PCG solves are $\geq$10x faster than the fastest QDLDL solve, and the slowest GBD-PCG solve is only 2.5x slower than the slowest QDLDL solve (with only 10\% of solves $\geq$2x slower).}
    \vspace{-12pt}
    \label{fig:solve-times-distribution}
\end{figure}
\fi

\textbf{Performance Distribution:} Importantly, in most cases, the speedup is much larger than this. This is because iterative methods have a variable runtime as they can exit early depending upon the exit tolerance, $\epsilon$. We demonstrate this using the 128 knot point problem as a case study in Figure~\ref{fig:solve-times-distribution}.

We plot the distribution of solve times for QDLDL against GPU-PCG $\epsilon = 1e^{-4}$, resulting in our 1.9x average speedup in Figure~\ref{fig:solve-times}, as well as $\epsilon = 5e^{-5}$ and $\epsilon = 1e^{-5}$. QDLDL presents a uni-modal timing distribution with almost all solves occurring between 280 and 305 $\mu$s. GBD-PCG, on the other hand, presents a bi-modal timing distribution clustered both much faster and a little slower than QDLDL. For example, for $\epsilon = 1e^{-4}$, 65\% of GBD-PCG solves are $\geq$10x faster than the fastest QDLDL solve, and the slowest GBD-PCG solve is only 2.5x slower than the slowest QDLDL solve (with only 10\% of solves $\geq$2x slower).

Furthermore, while all values of $\epsilon$ shown in the plot were able to successfully track the target trajectory, the lower the exit tolerance, the more of the distribution mass shifted to being $\geq$10x faster than QDLDL (65\%, 52\%, 20\% for $\epsilon = 1e^{-4}, 5e^{-5}, 1e^{-5}$ respectively). However, when $\epsilon$ was reduced even farther, our entire NMPC controller was unable to accurately track our target trajectory. These results present interesting directions for future work to 
find ways to eliminate the second slower mode of the solve time distribution while ensuring robust NMPC convergence.

\subsection{End-to-End NMPC Performance} \label{sec:results:MPC}
To validate efficacy for use in NMPC for robotics applications, we also demonstrate the impact of our approach on the number of iterations of MPCGPU we could achieve at each control step for varying control rates and trajectory lengths. Figure~\ref{fig:heatmap} shows the resulting number of average trajectory optimization solver iterations we can compute while meeting the specified control rates and trajectory lengths using both our GBD-PCG solver as well as QDLDL to solve the thousands of underlying linear systems.\footnote{We note that in the QDLDL case, as the NMPC loop is running on the GPU, data needs to be copied onto the host before executing the solve and converted into the sparse CSR matrix format. To ensure fair comparisons and avoid overheads for unnecessary data transfers and transformations, we implemented a variant of our parallel Schur complement computation which directly stores data in the CSR format expected by QDLDL.}

Regardless of the linear system solver, our GPU-first approach, with both fast parallel construction of the Schur complement and fast parallel computation of the line search, enables trajectories as long as $128$ knot points to operate at a $1$kHz control rate, and achieve at least $4$ iterations at a $500$Hz control rate, for a per-iteration rate of $2$kHz.
Furthermore, similar to what we witness in the case of average linear system solve times, as the problem gets larger and the control rate increases, our fully GPU-based approach is increasingly performant.
Highlights include our approach's ability to scale to $512$ knot points at a $1$kHz control rate and execute $8$ iterations for 128 knot points at a $500$Hz control rate, for a per-iteration rate of $4$kHz.
These results compare favorably to previously reported results in the literature of about $500$hz to $1$kHz per-iteration rates for trajectories of $30$ to $120$ knot points using state-of-the-art CPU-based~\cite{mastalli2020crocoddyl,kleff2021high} and GPU-based~\cite{Plancher18} solvers for similar NMPC tasks. 
As such, our GPU-first approach opens up the possibility for our NMPC solver to either leverage longer horizon trajectories, run at faster control rates, produce more optimal solutions for the same horizon and control rate, or include some combination of those highly beneficial traits.

\iffigs
\begin{figure}[!t]
    \centering
    \vspace{5pt}
    \includegraphics[width=0.99\columnwidth]{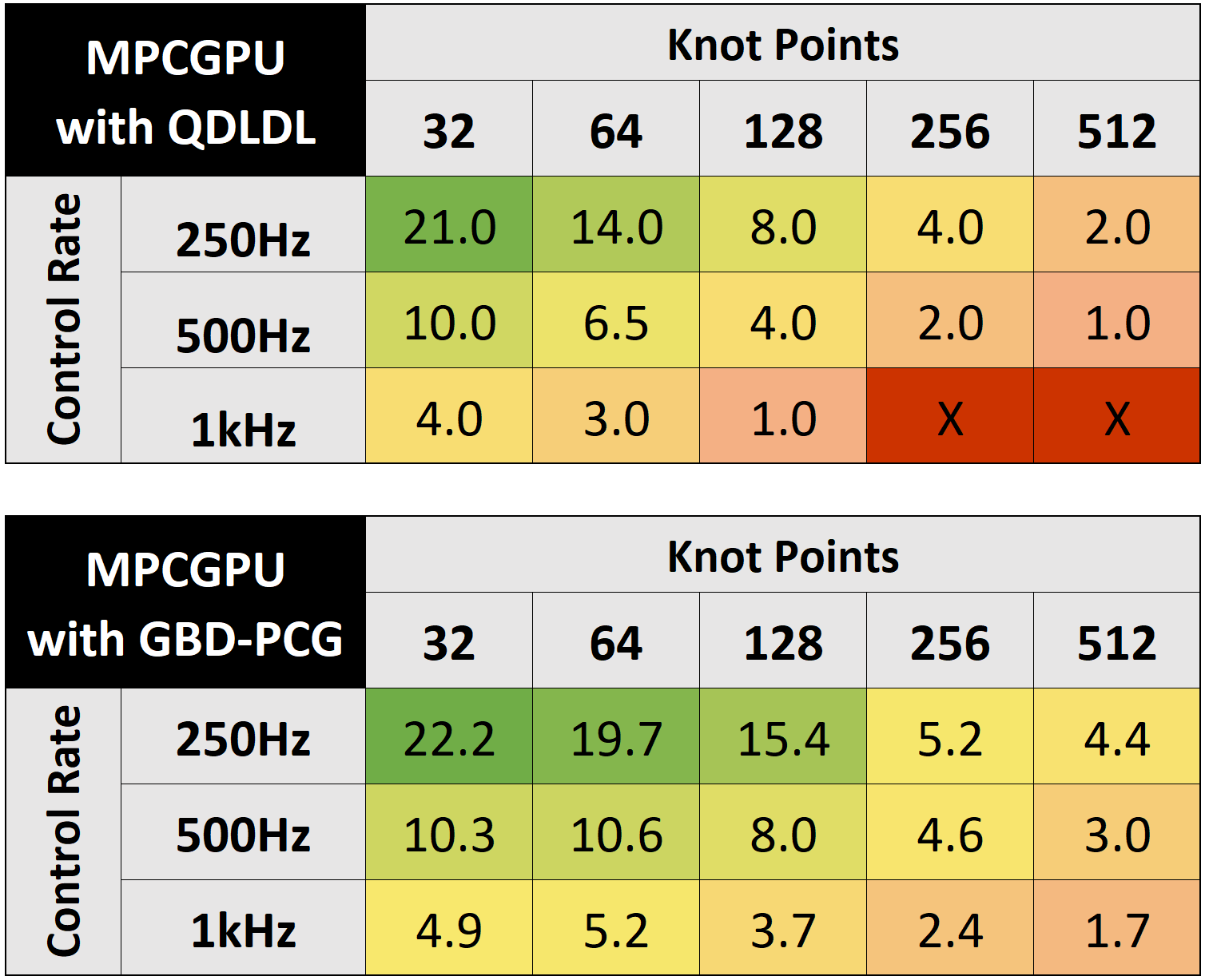}
    \vspace{-8pt}
    \caption{Average number of trajectory optimization iterations of MPCGPU executed at each control step for varying control rates and trajectory lengths. Our GPU-first approach to Schur complement construction and our parallel line search enables high control rates and long trajectories regardless of the underlying solver. However, the improved scalability of GBD-PCG enables our approach to scale to $512$ knot points at $1$kHz and execute $8$ iterations for 128 knot points at $500$Hz, for a per-iteration rate of $4$kHz.}
    \label{fig:heatmap}
    \vspace{-12pt}
\end{figure}
\fi

\section{Conclusion and Future Work} \label{sec:conclusion}
In this work, we introduce MPCGPU, a GPU-accelerated, real-time NMPC solver built around a parallel PCG solver. MPCGPU exploits the structured sparsity and natural parallelism in both direct trajectory optimization algorithms and iterative linear system solvers. Our experiments show that our approach is able to scale NMPC to larger problems, and operate it at faster rates, than is possible with existing state-of-the-art solvers. In particular, for tracking tasks using the Kuka IIWA manipulator, MPCGPU is able to scale to kilohertz control rates with trajectories as long as 512 knot points. For this problem, our GPU-based PCG solver outperforms a state-of-the-art CPU-based linear system solver by as much as 10x for a majority of solves and 3.6x on average. 

There are many promising directions for future work to improve the functionality and usability of our approach. Most importantly, like all iterative methods, our GPU-based PCG solver exhibits variability in its solve times. Future work which learns when to leverage iterative methods vs. factorization-based methods, or which learns dynamic values for hyperparameters to reduce the worst-case runtimes, without sacrificing overall NMPC robustness, would greatly improve average-case performance. 
Furthermore, it would be interesting to explore the performance implications of adding additional constraints either through expanding the KKT system, or through augmented Lagrangian or operator splitting methods~\cite{Nocedal06, Howell19, zhou2020accelerated}. 
Finally, we would like to evaluate our approach on physical robots at the edge using low-power GPU platforms such as the NVIDIA Jetson~\cite{ditty2022nvidia}.

\bibliographystyle{bib/IEEEtran_new}
\bibliography{bib/IEEEabrv,bib/agile,bib/a2r}

\ifdefined\DeclarePrefChars\DeclarePrefChars{'’-}\else\fi
\begin{thebibliography}{10}
\providecommand{\url}[1]{#1}
\csname url@rmstyle\endcsname
\providecommand{\newblock}{\relax}
\providecommand{\bibinfo}[2]{#2}
\providecommand\BIBentrySTDinterwordspacing{\spaceskip=0pt\relax}
\providecommand\BIBentryALTinterwordstretchfactor{4}
\providecommand\BIBentryALTinterwordspacing{\spaceskip=\fontdimen2\font plus
\BIBentryALTinterwordstretchfactor\fontdimen3\font minus \fontdimen4\font\relax}
\providecommand\BIBforeignlanguage[2]{{%
\expandafter\ifx\csname l@#1\endcsname\relax
\typeout{** WARNING: IEEEtran.bst: No hyphenation pattern has been}%
\typeout{** loaded for the language `#1'. Using the pattern for}%
\typeout{** the default language instead.}%
\else
\language=\csname l@#1\endcsname
\fi
#2}}

\bibitem{hogan2018reactive}
F.~R. Hogan, E.~R. Grau, and A.~Rodriguez, ``Reactive planar manipulation with convex hybrid mpc,'' in \emph{2018 IEEE International Conference on Robotics and Automation (ICRA)}.\hskip 1em plus 0.5em minus 0.4em\relax IEEE, 2018, pp. 247--253.

\bibitem{sleiman2021unified}
J.-P. Sleiman, F.~Farshidian, M.~V. Minniti, and M.~Hutter, ``A unified mpc framework for whole-body dynamic locomotion and manipulation,'' \emph{IEEE Robotics and Automation Letters}, vol.~6, no.~3, pp. 4688--4695, 2021.

\bibitem{tranzatto2022cerberus}
M.~Tranzatto, F.~Mascarich, L.~Bernreiter, C.~Godinho, M.~Camurri, S.~Khattak, T.~Dang, V.~Reijgwart, J.~Loeje, D.~Wisth, S.~Zimmermann, H.~Nguyen, M.~Fehr, L.~Solanka, R.~Buchanan, M.~Bjelonic, N.~Khedekar, M.~Valceschini, F.~Jenelten, M.~Dharmadhikari, T.~Homberger, P.~D. Petris, L.~Wellhausen, M.~Kulkarni, T.~Miki, S.~Hirsch, M.~Montenegro, C.~Papachristos, F.~Tresoldi, J.~Carius, G.~Valsecchi, J.~Lee, K.~Meyer, X.~Wu, J.~Nieto, A.~Smith, M.~Hutter, R.~Siegwart, M.~Mueller, M.~Fallon, and K.~Alexis, ``Cerberus: Autonomous legged and aerial robotic exploration in the tunnel and urban circuits of the darpa subterranean challenge,'' \emph{arXiv preprint arXiv:2201.07067}, 2022.

\bibitem{wensing2022optimization}
P.~M. Wensing, M.~Posa, Y.~Hu, A.~Escande, N.~Mansard, and A.~Del~Prete, ``Optimization-based control for dynamic legged robots,'' \emph{arXiv preprint arXiv:2211.11644}, 2022.

\bibitem{Kuindersma23Talk}
S.~Kuindersma, ``Taskable agility: Making useful dynamic behavior easier to create,'' Princeton Robotics Seminar, April 2023.

\bibitem{Betts01}
J.~T. Betts, \emph{Practical Methods for Optimal Control Using Nonlinear Programming}, ser. Advances in Design and Control.\hskip 1em plus 0.5em minus 0.4em\relax {Society for Industrial and Applied Mathematics (SIAM)}, vol.~3.

\bibitem{Bellman57}
R.~Bellman, \emph{Dynamic {{Programming}}}.\hskip 1em plus 0.5em minus 0.4em\relax {Dover}.

\bibitem{Mayne66}
D.~Q. Mayne, ``A second-order gradient method of optimizing non- linear discrete time systems,'' vol.~3, p. 8595.

\bibitem{Jacobson70}
D.~H. Jacobson and D.~Q. Mayne, ``Differential dynamic programming,'' 1970.

\bibitem{Nocedal06}
J.~Nocedal and S.~J. Wright, \emph{Numerical {{Optimization}}}, 2nd~ed.\hskip 1em plus 0.5em minus 0.4em\relax {Springer}.

\bibitem{Betts91}
J.~T. Betts and W.~P. Huffman, ``Trajectory optimization on a parallel processor,'' vol.~14, no.~2, pp. 431--439.

\bibitem{Esmaeilzadeh11}
H.~Esmaeilzadeh, E.~Blem, R.~St.~Amant, K.~Sankaralingam, and D.~Burger, ``Dark {{Silicon}} and the {{End}} of {{Multicore Scaling}},'' in \emph{Proceedings of the 38th {{Annual International Symposium}} on {{Computer Architecture}}}, ser. ISCA '11.\hskip 1em plus 0.5em minus 0.4em\relax {ACM}, pp. 365--376.

\bibitem{Venkatesh10}
G.~Venkatesh, J.~Sampson, N.~Goulding, S.~Garcia, V.~Bryksin, J.~Lugo-Martinez, S.~Swanson, and M.~B. Taylor, ``Conservation {{Cores}}: {{Reducing}} the {{Energy}} of {{Mature Computations}},'' in \emph{Proceedings of the {{Fifteenth Edition}} of {{ASPLOS}} on {{Architectural Support}} for {{Programming Languages}} and {{Operating Systems}}}, ser. ASPLOS XV.\hskip 1em plus 0.5em minus 0.4em\relax {ACM}, pp. 205--218.

\bibitem{Antony17}
T.~Antony and M.~J. Grant, ``Rapid {{Indirect Trajectory Optimization}} on {{Highly Parallel Computing Architectures}},'' vol.~54, no.~5, pp. 1081--1091.

\bibitem{Pan19}
Z.~Pan, B.~Ren, and D.~Manocha, ``Gpu-based contact-aware trajectory optimization using a smooth force model,'' in \emph{Proceedings of the 18th Annual ACM SIGGRAPH/Eurographics Symposium on Computer Animation}, ser. SCA '19.\hskip 1em plus 0.5em minus 0.4em\relax New York, NY, USA: ACM, 2019, pp. 4:1--4:12.

\bibitem{Plancher21}
B.~Plancher, S.~M. Neuman, T.~Bourgeat, S.~Kuindersma, S.~Devadas, and V.~J. Reddi, ``Accelerating robot dynamics gradients on a cpu, gpu, and fpga,'' \emph{IEEE Robotics and Automation Letters}, vol.~6, no.~2, pp. 2335--2342, 2021.

\bibitem{Neuman21}
\BIBentryALTinterwordspacing
S.~M. Neuman, B.~Plancher, T.~Bourgeat, T.~Tambe, S.~Devadas, and V.~J. Reddi, ``Robomorphic computing: A design methodology for domain-specific accelerators parameterized by robot morphology,'' ser. ASPLOS 2021.\hskip 1em plus 0.5em minus 0.4em\relax New York, NY, USA: Association for Computing Machinery, 2021, p. 674–686. [Online]. Available: \url{https://doi-org.ezp-prod1.hul.harvard.edu/10.1145/3445814.3446746}
\BIBentrySTDinterwordspacing

\bibitem{Plancher22GRiDGPUAcceleratedRigidBodyDynamicsAnalyticalGradientsb}
\BIBentryALTinterwordspacing
B.~Plancher, S.~M. Neuman, R.~Ghosal, S.~Kuindersma, and V.~J. Reddi, ``{{GRiD}}: {{GPU-Accelerated Rigid Body Dynamics}} with {{Analytical Gradients}},'' in \emph{2022 {{International Conference}} on {{Robotics}} and {{Automation}} ({{ICRA}})}.\hskip 1em plus 0.5em minus 0.4em\relax {IEEE}, pp. 6253--6260. [Online]. Available: \url{https://ieeexplore.ieee.org/document/9812384/}
\BIBentrySTDinterwordspacing

\bibitem{Lee22ParallelILQR}
Y.~Lee, M.~Cho, and K.-S. Kim, ``Gpu-parallelized iterative lqr with input constraints for fast collision avoidance of autonomous vehicles,'' in \emph{2022 IEEE/RSJ International Conference on Intelligent Robots and Systems (IROS)}, 2022, pp. 4797--4804.

\bibitem{Neuman23RoboShape}
\BIBentryALTinterwordspacing
S.~M. Neuman, R.~Ghosal, T.~Bourgeat, B.~Plancher, and V.~J. Reddi, ``Roboshape: Using topology patterns to scalably and flexibly deploy accelerators across robots,'' in \emph{Proceedings of the 50th Annual International Symposium on Computer Architecture}, ser. ISCA '23.\hskip 1em plus 0.5em minus 0.4em\relax New York, NY, USA: Association for Computing Machinery, 2023. [Online]. Available: \url{https://doi.org/10.1145/3579371.3589104}
\BIBentrySTDinterwordspacing

\bibitem{yang2023rbdcore}
Y.~Yang, X.~Chen, and Y.~Han, ``Rbdcore: Robot rigid body dynamics accelerator with multifunctional pipelines,'' \emph{arXiv preprint arXiv:2307.02274}, 2023.

\bibitem{Giftthaler17}
\BIBentryALTinterwordspacing
M.~Giftthaler, M.~Neunert, M.~Stäuble, J.~Buchli, and M.~Diehl, ``A {{Family}} of {{Iterative Gauss}}-{{Newton Shooting Methods}} for {{Nonlinear Optimal Control}}.'' [Online]. Available: \url{http://arxiv.org/abs/1711.11006}
\BIBentrySTDinterwordspacing

\bibitem{Farshidian17}
F.~Farshidian, E.~Jelavic, A.~Satapathy, M.~Giftthaler, and J.~Buchli, ``Real-time motion planning of legged robots: {{A}} model predictive control approach,'' in \emph{2017 {{IEEE}}-{{RAS}} 17th {{International Conference}} on {{Humanoid Robotics}}}.

\bibitem{Kouzoupis16}
D.~Kouzoupis, R.~Quirynen, B.~Houska, and M.~Diehl, ``A {{Block Based ALADIN Scheme}} for {{Highly Parallelizable Direct Optimal Control}},'' in \emph{Proceedings of the {{American Control Conference}}}.

\bibitem{jiang2020parallel}
Y.~Jiang, J.~Oravec, B.~Houska, and M.~Kvasnica, ``Parallel mpc for linear systems with input constraints,'' \emph{IEEE Transactions on Automatic Control}, vol.~66, no.~7, pp. 3401--3408, 2020.

\bibitem{Plancher18}
B.~Plancher and S.~Kuindersma, ``A {Performance} {Analysis} of {Parallel} {Differential} {Dynamic} {Programming} on a {GPU},'' in \emph{International Workshop on the Algorithmic Foundations of Robotics (WAFR)}.

\bibitem{Plancher19a}
------, ``Realtime model predictive control using parallel ddp on a gpu,'' in \emph{Toward Online Optimal Control of Dynamic Robots Workshop at the 2019 International Conference on Robotics and Automation (ICRA)}, Montreal, Canada, May. 2019.

\bibitem{eisenstat1981efficient}
S.~C. Eisenstat, ``Efficient implementation of a class of preconditioned conjugate gradient methods,'' \emph{SIAM Journal on Scientific and Statistical Computing}, vol.~2, no.~1, pp. 1--4, 1981.

\bibitem{Saad03}
Y.~Saad, \emph{Iterative methods for sparse linear systems}.\hskip 1em plus 0.5em minus 0.4em\relax SIAM, 2003.

\bibitem{Plancher22Dissertation}
B.~Plancher, ``Gpu acceleration for real-time, whole-body, nonlinear model predictive control,'' Ph.D. dissertation, Harvard University, Cambridge, MA, USA, April 2022.

\bibitem{Helfenstein12ParallelpreconditionedconjugategradientalgorithmGPU}
\BIBentryALTinterwordspacing
R.~Helfenstein and J.~Koko, ``Parallel preconditioned conjugate gradient algorithm on {{GPU}},'' vol. 236, no.~15, pp. 3584--3590. [Online]. Available: \url{https://www.sciencedirect.com/science/article/pii/S0377042711002196}
\BIBentrySTDinterwordspacing

\bibitem{Schubiger20}
M.~Schubiger, G.~Banjac, and J.~Lygeros, ``Gpu acceleration of admm for large-scale quadratic programming,'' \emph{Journal of Parallel and Distributed Computing}, vol. 144, pp. 55--67, 2020.

\bibitem{Jung06}
J.~H. Jung and D.~P. O’Leary, ``Cholesky decomposition and linear programming on a gpu,'' \emph{Scholarly Paper, University of Maryland}, 2006.

\bibitem{Yang12}
D.~Yang, G.~D. Peterson, and H.~Li, ``Compressed sensing and cholesky decomposition on fpgas and gpus,'' \emph{Parallel Computing}, vol.~38, no.~8, pp. 421--437, 2012.

\bibitem{Venetis15}
I.~E. Venetis, A.~Kouris, A.~Sobczyk, E.~Gallopoulos, and A.~H. Sameh, ``A direct tridiagonal solver based on givens rotations for gpu architectures,'' \emph{Parallel Computing}, vol.~49, pp. 101--116, 2015.

\bibitem{Hogg16}
J.~D. Hogg, E.~Ovtchinnikov, and J.~A. Scott, ``A sparse symmetric indefinite direct solver for gpu architectures,'' \emph{ACM Transactions on Mathematical Software (TOMS)}, vol.~42, no.~1, pp. 1--25, 2016.

\bibitem{Hu17}
X.~Hu, C.~C. Douglas, R.~Lumley, and M.~Seo, ``Gpu accelerated sequential quadratic programming,'' in \emph{2017 16th International Symposium on Distributed Computing and Applications to Business, Engineering and Science (DCABES)}.\hskip 1em plus 0.5em minus 0.4em\relax IEEE, 2017, pp. 3--6.

\bibitem{Yeralan17}
S.~N. Yeralan, T.~A. Davis, W.~M. Sid-Lakhdar, and S.~Ranka, ``Algorithm 980: Sparse qr factorization on the gpu,'' \emph{ACM Transactions on Mathematical Software (TOMS)}, vol.~44, no.~2, pp. 1--29, 2017.

\bibitem{swirydowicz2022linear}
K.~{\'S}wirydowicz, E.~Darve, W.~Jones, J.~Maack, S.~Regev, M.~A. Saunders, S.~J. Thomas, and S.~Pele{\v{s}}, ``Linear solvers for power grid optimization problems: a review of gpu-accelerated linear solvers,'' \emph{Parallel Computing}, vol. 111, p. 102870, 2022.

\bibitem{cole2023exploiting}
D.~Cole, S.~Shin, F.~Pacaud, V.~M. Zavala, and M.~Anitescu, ``Exploiting gpu/simd architectures for solving linear-quadratic mpc problems,'' in \emph{2023 American Control Conference (ACC)}.\hskip 1em plus 0.5em minus 0.4em\relax IEEE, 2023, pp. 3995--4000.

\bibitem{shin2023accelerating}
S.~Shin, F.~Pacaud, and M.~Anitescu, ``Accelerating optimal power flow with gpus: Simd abstraction of nonlinear programs and condensed-space interior-point methods,'' \emph{arXiv preprint arXiv:2307.16830}, 2023.

\bibitem{pacaud2023accelerating}
F.~Pacaud, S.~Shin, M.~Schanen, D.~A. Maldonado, and M.~Anitescu, ``Accelerating condensed interior-point methods on simd/gpu architectures,'' \emph{Journal of Optimization Theory and Applications}, pp. 1--20, 2023.

\bibitem{Bolz03}
\BIBentryALTinterwordspacing
J.~Bolz, I.~Farmer, E.~Grinspun, and P.~Schröoder, ``Sparse {{Matrix Solvers}} on the {{GPU}}: {{Conjugate Gradients}} and {{Multigrid}},'' in \emph{{{ACM SIGGRAPH}} 2003 {{Papers}}}, ser. SIGGRAPH '03.\hskip 1em plus 0.5em minus 0.4em\relax {ACM}, pp. 917--924. [Online]. Available: \url{http://doi.acm.org/10.1145/1201775.882364}
\BIBentrySTDinterwordspacing

\bibitem{Liu13}
H.~Liu, J.-H. Seo, R.~Mittal, and H.~H. Huang, ``Gpu-accelerated scalable solver for banded linear systems,'' in \emph{2013 IEEE International Conference on Cluster Computing (CLUSTER)}.\hskip 1em plus 0.5em minus 0.4em\relax IEEE, 2013, pp. 1--8.

\bibitem{Anzt17}
H.~Anzt, M.~Gates, J.~Dongarra, M.~Kreutzer, G.~Wellein, and M.~Köhler, ``Preconditioned krylov solvers on gpus,'' \emph{Parallel Computing}, 05 2017.

\bibitem{Anzt18}
H.~Anzt, M.~Kreutzer, E.~Ponce, G.~D. Peterson, G.~Wellein, and J.~Dongarra, ``Optimization and performance evaluation of the idr iterative krylov solver on gpus,'' \emph{The International Journal of High Performance Computing Applications}, vol.~32, no.~2, pp. 220--230, 2018.

\bibitem{Flegar19}
G.~Flegar \emph{et~al.}, ``Sparse linear system solvers on gpus: Parallel preconditioning, workload balancing, and communication reduction,'' Ph.D. dissertation, Universitat Jaume I, 2019.

\bibitem{tiwari2022strategies}
M.~Tiwari and S.~Vadhiyar, ``Strategies for efficient execution of pipelined conjugate gradient method on gpu systems,'' in \emph{International Conference on High Performance Computing}.\hskip 1em plus 0.5em minus 0.4em\relax Springer, 2022, pp. 77--89.

\bibitem{Chang12}
L.-W. Chang, J.~A. Stratton, H.-S. Kim, and W.-M.~W. Hwu, ``A scalable, numerically stable, high-performance tridiagonal solver using gpus,'' in \emph{SC'12: Proceedings of the International Conference on High Performance Computing, Networking, Storage and Analysis}.\hskip 1em plus 0.5em minus 0.4em\relax IEEE, 2012, pp. 1--11.

\bibitem{Chang14}
\BIBentryALTinterwordspacing
L.-W. Chang and W.-m.~W. Hwu, \emph{A Guide for Implementing Tridiagonal Solvers on GPUs}.\hskip 1em plus 0.5em minus 0.4em\relax Springer International Publishing, 2014, pp. 29--44. [Online]. Available: \url{https://doi.org/10.1007/978-3-319-06548-9_2}
\BIBentrySTDinterwordspacing

\bibitem{Dieguez15}
A.~P. Di{\'e}guez, M.~Amor, and R.~Doallo, ``New tridiagonal systems solvers on gpu architectures,'' in \emph{2015 IEEE 22nd International Conference on High Performance Computing (HiPC)}.\hskip 1em plus 0.5em minus 0.4em\relax IEEE, 2015, pp. 85--94.

\bibitem{Lamas18}
A.~Lamas~Daviña and J.~Roman, ``Mpi-cuda parallel linear solvers for block-tridiagonal matrices in the context of slepc’s eigensolvers,'' \emph{Parallel computing}, vol.~74, pp. 118--135, 2018.

\bibitem{naumov2011incomplete}
M.~Naumov, ``Incomplete-lu and cholesky preconditioned iterative methods using cusparse and cublas,'' \emph{Nvidia white paper}, vol.~3, 2011.

\bibitem{Heinrich15}
S.~Heinrich, A.~Zoufahl, and R.~Rojas, ``Real-time trajectory optimization under motion uncertainty using a {{GPU}},'' in \emph{2015 {{IEEE}}/{{RSJ International Conference}} on {{Intelligent Robots}} and {{Systems}} ({{IROS}})}, pp. 3572--3577.

\bibitem{Wu16}
Q.~Wu, F.~Xiong, F.~Wang, and Y.~Xiong, ``Parallel particle swarm optimization on a graphics processing unit with application to trajectory optimization,'' \emph{Engineering Optimization}, vol.~48, no.~10, pp. 1679--1692, 2016.

\bibitem{Williams17}
G.~Williams, A.~Aldrich, and E.~A. Theodorou, ``Model predictive path integral control: From theory to parallel computation,'' \emph{Journal of Guidance, Control, and Dynamics}, vol.~40, no.~2, pp. 344--357, 2017.

\bibitem{Phung17}
\BIBentryALTinterwordspacing
D.-K. Phung, B.~Hérissé, J.~Marzat, and S.~Bertrand, ``Model {{Predictive Control}} for {{Autonomous Navigation Using Embedded Graphics Processing Unit}},'' vol.~50, no.~1, pp. 11\,883--11\,888. [Online]. Available: \url{http://www.sciencedirect.com/science/article/pii/S2405896317319614}
\BIBentrySTDinterwordspacing

\bibitem{Hyatt17}
P.~Hyatt and M.~D. Killpack, ``Real-time evolutionary model predictive control using a graphics processing unit,'' in \emph{2017 IEEE-RAS 17th International Conference on Humanoid Robotics (Humanoids)}.\hskip 1em plus 0.5em minus 0.4em\relax IEEE, 2017, pp. 569--576.

\bibitem{Ohyama17}
S.~Ohyama and H.~Date, ``Parallelized nonlinear model predictive control on gpu,'' in \emph{2017 11th Asian Control Conference (ASCC)}.\hskip 1em plus 0.5em minus 0.4em\relax IEEE, 2017, pp. 1620--1625.

\bibitem{Rathai19}
K.~M.~M. Rathai, O.~Sename, and M.~Alamir, ``Gpu-based parameterized nmpc scheme for control of half car vehicle with semi-active suspension system,'' \emph{IEEE Control Systems Letters}, vol.~3, no.~3, pp. 631--636, 2019.

\bibitem{Wang19a}
\BIBentryALTinterwordspacing
Y.~Wang, X.~Luo, F.~Zhang, and S.~Wang, ``Gpu-based model predictive control for continuous casting spray cooling control system using particle swarm optimization,'' \emph{Control Engineering Practice}, vol.~84, pp. 349--364, 2019. [Online]. Available: \url{https://www.sciencedirect.com/science/article/pii/S096706611830710X}
\BIBentrySTDinterwordspacing

\bibitem{Hyatt20}
P.~Hyatt, C.~S. Williams, and M.~D. Killpack, ``Parameterized and gpu-parallelized real-time model predictive control for high degree of freedom robots,'' \emph{arXiv preprint arXiv:2001.04931}, 2020.

\bibitem{Frasch13}
J.~V. Frasch, M.~Vukov, H.~J. Ferreau, and M.~Diehl, ``A dual newton strategy for the efficient solution of sparse quadratic programs arising in sqp-based nonlinear mpc,'' \emph{Optimization Online 3972}, 2013.

\bibitem{Astudillo22}
A.~Astudillo, J.~Gillis, G.~Pipeleers, W.~Decré, and J.~Swevers, ``Speed-up of nonlinear model predictive control for robot manipulators using task and data parallelism,'' in \emph{2022 IEEE 17th International Conference on Advanced Motion Control (AMC)}, 2022, pp. 201--206.

\bibitem{Gang12}
Y.~Gang and L.~Mingguang, ``Acceleration of mpc using graphic processing unit,'' in \emph{Proceedings of 2012 2nd International Conference on Computer Science and Network Technology}.\hskip 1em plus 0.5em minus 0.4em\relax IEEE, 2012, pp. 1001--1004.

\bibitem{Gade12}
N.~F. Gade-Nielsen, J.~B. J{\o}rgensen, and B.~Dammann, ``Mpc toolbox with gpu accelerated optimization algorithms,'' in \emph{10th European workshop on advanced control and diagnosis}.\hskip 1em plus 0.5em minus 0.4em\relax Technical University of Denmark, 2012.

\bibitem{Huang11}
Y.~Huang, K.~V. Ling, and S.~See, ``Solving {{Quadratic Programming Problems}} on {{Graphics Processing Unit}}.''

\bibitem{Yu17}
\BIBentryALTinterwordspacing
L.~Yu, A.~Goldsmith, and S.~Di~Cairano, ``Efficient {{Convex Optimization}} on {{GPUs}} for {{Embedded Model Predictive Control}},'' in \emph{Proceedings of the {{General Purpose GPUs}}}, ser. GPGPU-10.\hskip 1em plus 0.5em minus 0.4em\relax {ACM}, pp. 12--21. [Online]. Available: \url{http://doi.acm.org/10.1145/3038228.3038234}
\BIBentrySTDinterwordspacing

\bibitem{Featherstone08}
R.~Featherstone, \emph{Rigid {{Body Dynamics Algorithms}}}.\hskip 1em plus 0.5em minus 0.4em\relax {Springer}.

\bibitem{Wachter06}
\BIBentryALTinterwordspacing
A.~Wächter and L.~T. Biegler, ``On the {{Implementation}} of an {{Interior}}-point {{Filter Line}}-search {{Algorithm}} for {{Large}}-scale {{Nonlinear Programming}},'' vol. 106, no.~1, pp. 25--57. [Online]. Available: \url{https://doi.org/10.1007/s10107-004-0559-y}
\BIBentrySTDinterwordspacing

\bibitem{Gill05}
P.~E. Gill, W.~Murray, and M.~A. Saunders, ``{{SNOPT}}: {{An SQP Algorithm}} for {{Large}}-scale {{Constrained Optimization}},'' vol.~47, no.~1, pp. 99--131.

\bibitem{Schubiger20GPUAccelerationADMMLargeScaleQuadraticProgramming}
\BIBentryALTinterwordspacing
M.~Schubiger, G.~Banjac, and J.~Lygeros, ``{{GPU Acceleration}} of {{ADMM}} for {{Large-Scale Quadratic Programming}},'' vol. 144, pp. 55--67. [Online]. Available: \url{http://arxiv.org/abs/1912.04263}
\BIBentrySTDinterwordspacing

\bibitem{Nickolls08ScalableParallelProgrammingCUDA}
J.~Nickolls, I.~Buck, M.~Garland, and K.~Skadron, ``Scalable {{Parallel Programming}} with {{CUDA}},'' vol.~6, pp. 40--53.

\bibitem{NVIDIA18}
\BIBentryALTinterwordspacing
NVIDIA, \emph{{{NVIDIA CUDA C Programming Guide}}}, version 9.1~ed. [Online]. Available: \url{http://docs.nvidia.com/cuda/pdf/CUDA_C_Programming_Guide.pdf}
\BIBentrySTDinterwordspacing

\bibitem{bu2024symmetric}
X.~Bu and B.~Plancher, ``Symmetric stair preconditioning of linear systems for parallel trajectory optimization,'' in \emph{2024 IEEE International Conference on Robotics and Automation (ICRA)}.\hskip 1em plus 0.5em minus 0.4em\relax IEEE, 2024, pp. 9779--9786.

\bibitem{stellato20osqp}
\BIBentryALTinterwordspacing
B.~Stellato, G.~Banjac, P.~Goulart, A.~Bemporad, and S.~Boyd, ``{OSQP}: an operator splitting solver for quadratic programs,'' \emph{Mathematical Programming Computation}, vol.~12, no.~4, pp. 637--672, 2020. [Online]. Available: \url{https://doi.org/10.1007/s12532-020-00179-2}
\BIBentrySTDinterwordspacing

\bibitem{mastalli2020crocoddyl}
C.~Mastalli, R.~Budhiraja, W.~Merkt, G.~Saurel, B.~Hammoud, M.~Naveau, J.~Carpentier, L.~Righetti, S.~Vijayakumar, and N.~Mansard, ``Crocoddyl: An efficient and versatile framework for multi-contact optimal control,'' in \emph{2020 IEEE International Conference on Robotics and Automation (ICRA)}.\hskip 1em plus 0.5em minus 0.4em\relax IEEE, 2020, pp. 2536--2542.

\bibitem{kleff2021high}
S.~Kleff, A.~Meduri, R.~Budhiraja, N.~Mansard, and L.~Righetti, ``High-frequency nonlinear model predictive control of a manipulator,'' in \emph{2021 IEEE International Conference on Robotics and Automation (ICRA)}.\hskip 1em plus 0.5em minus 0.4em\relax IEEE, 2021, pp. 7330--7336.

\bibitem{Howell19}
T.~Howell, B.~Jackson, and Z.~Manchester, ``Altro: A fast solver for constrained trajectory optimization,'' in \emph{Proceedings of (IROS) IEEE/RSJ International Conference on Intelligent Robots and Systems}, November 2019, pp. 7674 -- 7679.

\bibitem{zhou2020accelerated}
Z.~Zhou and Y.~Zhao, ``Accelerated admm based trajectory optimization for legged locomotion with coupled rigid body dynamics.''\hskip 1em plus 0.5em minus 0.4em\relax IEEE, 2020, pp. 5082--5089.

\bibitem{ditty2022nvidia}
M.~Ditty, ``Nvidia orin system-on-chip,'' in \emph{2022 IEEE Hot Chips 34 Symposium (HCS)}.\hskip 1em plus 0.5em minus 0.4em\relax IEEE Computer Society, 2022, pp. 1--17.

\end{thebibliography}

\end{document}